\documentclass{article}
\usepackage[preprint]{spconf}
\usepackage{amsmath,graphicx}
\usepackage{color}
\usepackage{amsfonts}
\usepackage{amssymb} 
\usepackage[font=small]{caption} 
\usepackage{comment}
\usepackage{tabularx}
\usepackage{doi}
\usepackage{graphicx} 
\usepackage{listings} 
\usepackage{lineno} 
\usepackage{physics} 
\usepackage{tikz} 
\usepackage{url} 
\usepackage{siunitx} 
\usepackage{subcaption} 
\usepackage{tcolorbox}
\usepackage{varioref} 
\usepackage{cleveref} 
\usepackage[normalem]{ulem}
\useunder{\uline}{\ul}{}
\usepackage{multirow}
\usepackage[numbers]{natbib}
\usepackage{enumitem}

\copyrightnotice{\copyright\ IEEE 2020}
\toappear{To appear in {\it Proc.\ ICASSP2020, May 04-08, 2020, Barcelona, Spain}}

\newlist{RQ}{enumerate}{1}
\setlist[RQ]{label=\textbf{RQ\,\arabic*},ref={RQ\,\arabic*}}

\title{Improving automated segmentation of radio shows\\with audio embeddings}

\name{Oberon Berlage$^{\dagger}$\sthanks{Work performed during an internship at FD Mediagroep.} 
	  \qquad 
	  Klaus-Michael Lux$^{\ddagger}$\footnotemark[1] 
	  \qquad 
	  David Graus$^{\mathsection}$}
	  \address{$^{\dagger}$ University of Amsterdam \\
			   $^{\ddagger}$ Radboud Universiteit Nijmegen \\
			   $^{\mathsection}$ FD Mediagroep}

\begin{document}
\ninept
\maketitle
\begin{abstract}
Audio features have been proven useful for increasing the performance of automated topic segmentation systems. 
This study explores the novel task of using audio embeddings for automated, topically coherent segmentation of radio shows.
We created three different audio embedding generators using multi-class classification tasks on three datasets from different domains.
We evaluate topic segmentation performance of the audio embeddings and compare it against a text-only baseline.
We find that a set-up including audio embeddings generated through a non-speech sound event classification task significantly outperforms our text-only baseline by 32.3\% in F1-measure. 
In addition, we find that different classification tasks yield audio embeddings that vary in segmentation performance.
\end{abstract}

\section{Introduction}

With the rise of on-demand audio such as podcasts, online radio archives, and audiobooks, structuring and processing long audio streams is becoming a more important task.
In particular, segmenting longer audio into smaller, topically coherent segments can be helpful in content serving and consumption~\cite{sappelli_smart_2018}.

The task of topic segmentation is defined by Purver as \emph{``automatically dividing single long recordings or transcripts into shorter, topically coherent segments''}~\cite{purver_topic_2011}. 
Common methods in topic segmentation mostly revolve around using ``lexical similarity'' (e.g. TextTiling by Hearst~\cite{hearst_texttiling:_1997}) or using ``distinctive boundary features'' (e.g. detecting stopwords or signalling words~\cite{maybury_discourse_1998}).

Prior research shows that incorporating information from the audio signal in topic segmentation methods can be beneficial~\cite{purver_topic_2011, tur_integrating_2001, tsunoo_hierarchical_2017}. 
Processing and representing audio is commonly done using hand-crafted features~\cite{dieleman_end--end_2014} or spectrogram representations of the audio signal~\cite{virtanen_machine_2018}.
With the rise of deep learning, it is now possible to yield feature representations from neural networks. 
A well known example of this would be word embeddings. 
Several studies explored the use of audio embeddings in tasks such as
automatic speech classification~\cite{bengio_word_2014,chung_audio_2016,settle_discriminative_2016}, 
text-based sound retrieval~\cite{vijayakumar_sound-word2vec:_2017} or 
sound event classification~\cite{kiela_learning_2017, aytar_soundnet:_2016,hershey_cnn_2016}.
To the best of our knowledge, audio embeddings for topic segmentation have not been studied before.

In this work we study the use of audio embeddings to generate a feature representation for improving topically coherent segmentation of radio shows. 

First, we generate audio embedding spaces by training models on a multi-class classification task.
Specifically, we use datasets from three different domains:
a non-speech sound event classification task, 
a part of radio fragment classification task, and 
a word classification task.

Then, we evaluate and compare the three audio embeddings using an existing topic segmentation approach~\cite{sheikh_topic_2017} and find that using audio embeddings generated from a sound event classification task significantly outperform a text-only baseline by 32.3\%. 
The combination of text features and audio embeddings in the segmentation task leads in all cases to an increase in recall and overall performance. 
We show that segmentation based on audio embeddings alone can have a comparable performance to a text-only system. 
Finally, we see a clear difference in the segmentation performance between the different embedding generators.

In this paper we answer the following research questions:
\begin{RQ}
  \item Can audio embeddings be used to improve automatic, topically coherent segmentation of radio shows?\label{rq1}
  \item What is the effect of different training tasks for embedding generators on the topic segmentation performance?\label{rq2}
\end{RQ}
\section{Audio Embeddings for Topic Segmentation}
In this paper we study the use of audio embeddings to represent audio for topic segmentation. 
When using audio in machine learning tasks, it is common to do feature extraction to represent the audio signal in a more compact and information-dense way~\cite{virtanen_machine_2018}. 
Hand-crafting features from audio signals is a time-consuming effort and may require knowledge that is very specific to the problem at hand~\cite{dieleman_end--end_2014}. 
Spectrograms are commonly used audio representations, as feature learning from raw audio has been shown to be inferior to feature learning from spectrograms~\cite{dieleman_end--end_2014,wyse_audio_2017} and spectrograms provide a good trade-off between dimensionality and information~\cite{wyse_audio_2017}.

Earlier work has shown that using acoustic information, such as duration of pauses or pitch, improves the performance of topic segmentation~\cite{tur_integrating_2001,tsunoo_hierarchical_2017,purver_topic_2011,hirschberg_acoustic_1998}.
However, using audio embeddings to represent audio for topic segmentation has, to the best of our knowledge, not been studied before. 
Since audio embeddings may capture higher level structures in compact representations, they offer an attractive advantage over raw or hand-crafted acoustic features. 

\begin{figure}[t!]
  \centering
 \includegraphics[width=0.75\linewidth]{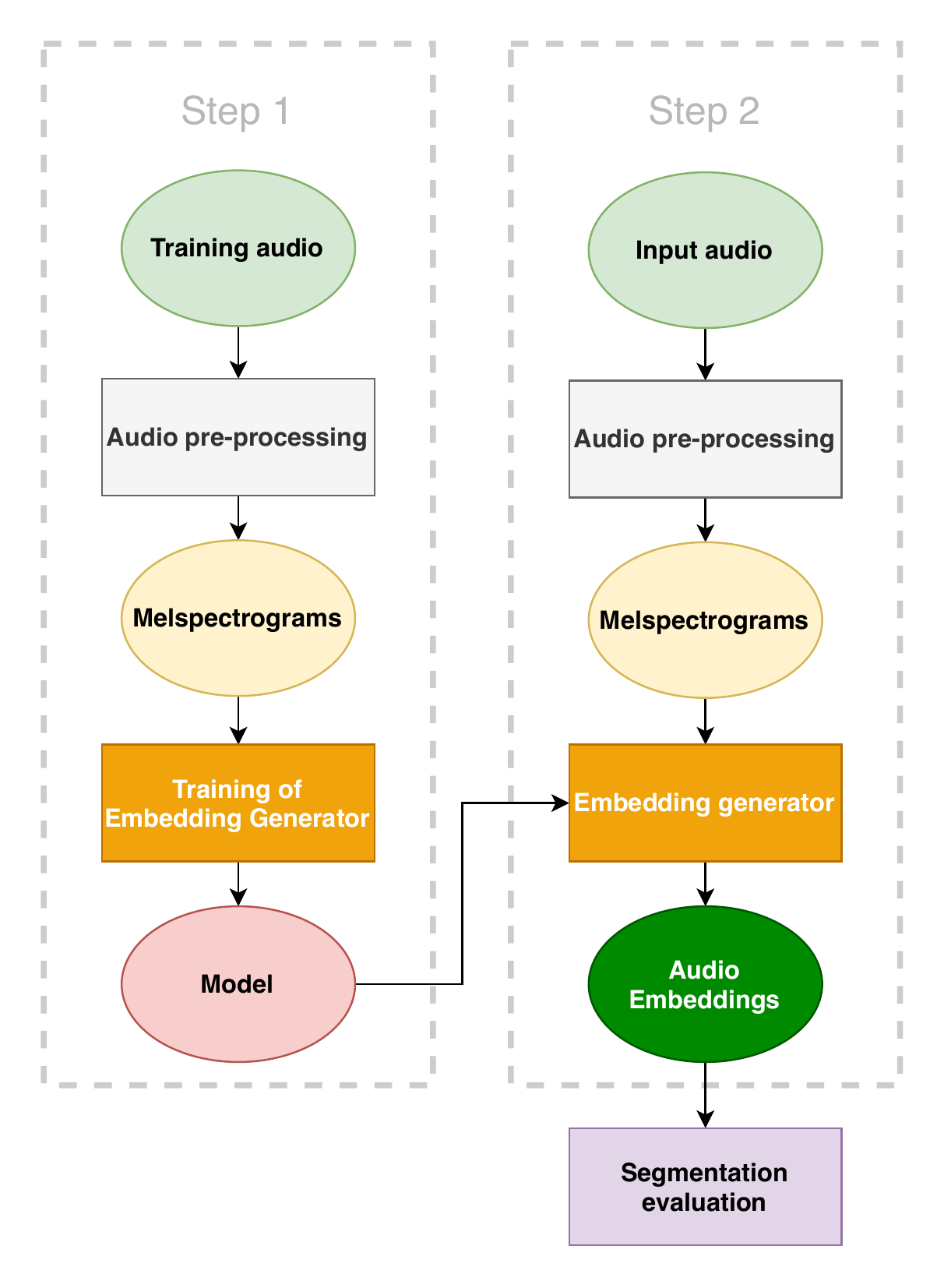}
 \caption{Overview of our experimental setup. 
 The left part depicts the training of our embedding generator. 
 The right part shows the generation of audio embeddings from input audio.}
 \label{fig:synthesis_overview}
\end{figure}

We generate audio embeddings in two steps, based on the procedures used by Bengio~\cite{bengio_word_2014} and Kiela~\cite{kiela_learning_2017}.
First, we train an ``embedding generator'' using a specific training task. 
Next, we apply the generator to the input audio of a down-stream task to obtain the audio embeddings.
This process is depicted in figure~\ref{fig:synthesis_overview}.

We obtain ``embedding generators'' by training Convolutional Neural Networks (CNNs) on different classification tasks. 
Different training tasks steer networks to yield different internal feature representations, by leveraging the discriminative nature of the classification task.
The network needs to learn a feature representation that best captures the distinctive properties between classes. 
Datasets vary in their class distinctions, and thus lead to different internal feature representations in the embedding generators that are trained. 
For example, a chirping bird and a thunderstorm can be distinguished by features related to the frequency of the sound. 
On the other hand, the sound of a chirping bird and a siren might both consist of higher tones, but their patterns over time are different. 
When learning to distinguish between spoken words, patterns in intensity and frequency might be more relevant features to learn.

\subsection{Classification tasks}
In this work, we train three audio embedding generators.
Our first embedding generator is trained on a Sound Event Classification (\textbf{SEC}) task. 
By learning to classify a diverse set of sound events, we expect the learned representation to be able to capture a more diverse set of sound properties.
Furthermore, as this task is not related to speech or radio shows, we expect it to learn a representation that is focused on non-speech sound properties.

The second embedding generator is trained on Fragment Part Classification (\textbf{FPC}), i.e., trained to classify different parts of radio fragments, such as the start, middle, end, and jingles. 
This approach is based on the observation that it is possible to hear a difference between starts and ends of radio fragments. 
A representation that is able to distinguish different parts of fragments might contain information that can be used as boundary features.

The third embedding generator is trained on a Word Classification (\textbf{WC}) or speech-to-text task. 
We expect this to provide us a representation of distinctive properties from human speech. 
By having to distinguish between different words, we expect the model to learn, e.g., differences in tones or word lengths.

\begin{figure}
  \centering
  \includegraphics[trim={0 .5cm 0 0},clip,width=0.70\linewidth]{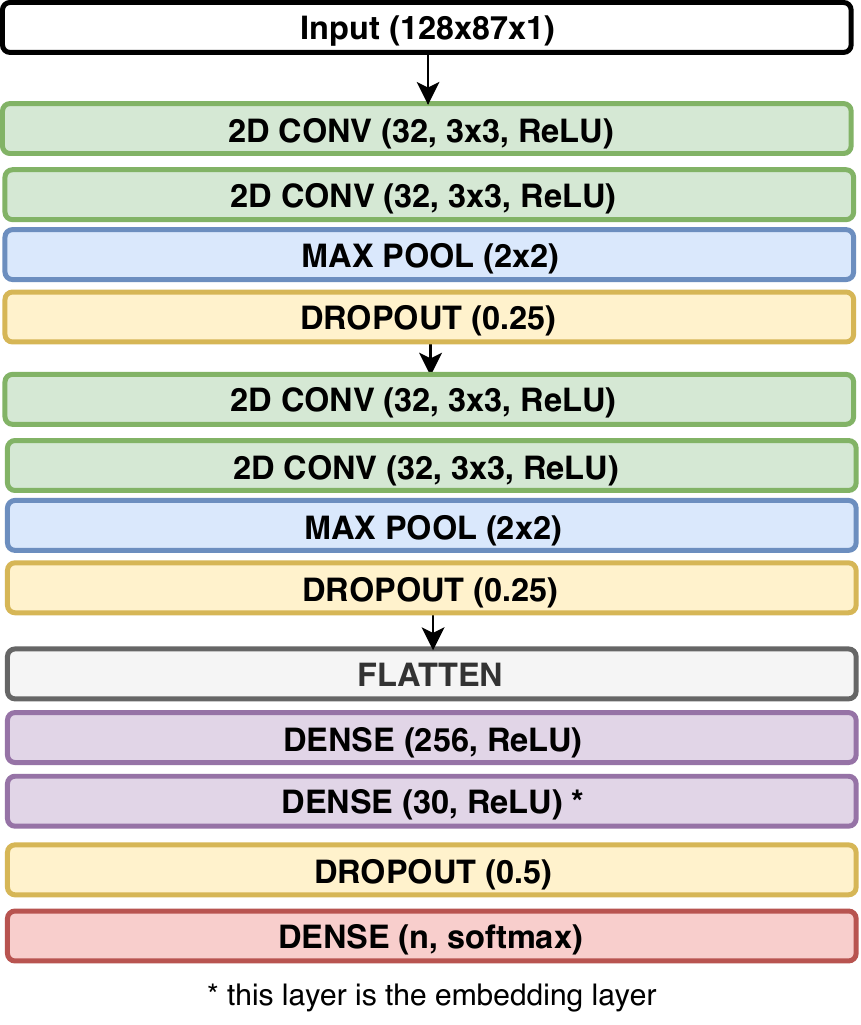}
  \vspace{-.25cm}
  \caption{An overview of the 2D Convolutional Neural Network architecture used for embedding generators. 
  The network takes Mel-scaled spectrogram as input. 
  The layer denoted with a $*$ is used as the embedding layer. 
  The final fully connected dense layer differs per task, with $n$ being the number of classes in the dataset.}
  \label{fig:vgg16network}
\end{figure}
\vspace{-.25cm}

\section{Experimental Setup}

\subsection{Classification datasets for embedding generators}
\label{sect:meth-datasets}

\begin{itemize}
\item[\textbf{SEC}]
Our first dataset is the sound events classification dataset ESC-50~\cite{piczak_esc:_2015}.
It consists of 50 classes of sound events, which consist of 40 audioclips of 5 seconds each.
The samples are manually annotated after retrieval from FreeSound\footnote{\url{https://freesound.org}}. 
They have been categorized into 5 major categories: 
Animals (e.g. dog barks), 
Natural soundscapes \& water (e.g. thunderstorm), 
Human (non-speech) (e.g. laughing), 
Interior/domestic (e.g. clock alarm), and 
Exterior/urban noises (e.g. sirens). 
We split all clips into 1 second chunks for further processing. 
\item[\textbf{FPC}]
Our second dataset is based on BNR Nieuwsradio\footnote{\url{https://bnr.nl/} (A news radio station part of FD Mediagroep)} audioclips and has four classes: three from different parts of radio fragments (begin, middle, end) and one from radio jingles
We extracted a random sample of 2500 audioclips with a duration between 2-15 minutes. 
Of these audioclips, we truncated the first and last 3 seconds, as they typically contained jingles. 
From the resulting fragments we cut the first and last 4 seconds, and randomly sampled 4 seconds from the remaining audio. 
We added samples from jingles as a fourth class, yielding 12,000 samples of 1-second audioclips in total.
\item[\textbf{WC}]
Our third dataset is constructed from spoken words in our corpus. 
The full dataset consists of 76,391 unique words and corresponding audioclips from BNR Nieuwsradio, from which we took a random subset, yielding 300 unique words with 100 samples each. 
We zero-padded all audioclips to 1 second. 
\end{itemize}

\subsection{Sound pre-processing}
\label{sect:meth-preproc}
All audio data was loaded as mono waveforms with a sample rate of 44.1kHz.
We downsampled clips with higher sample rates to 44.1kHz. 
All 1-second audioclips were turned into Mel-scaled spectrograms, resulting in matrices of 128 Mel-bands by 87 timesteps.

\subsection{Audio embedding training \& generation}
\label{sect:meth-traingen}
We generate audio embeddings in two stages: 
first, we train a CNN to yield an embedding generator, 
next, we generate audio embeddings from Mel-scaled spectrograms of our input audio. 

We explored different CNN architectures based on image classification CNNs for our embedding generators, using a sample of the audio classification task. 
We found that a downscaled version of the original VGG16 network~\cite{simonyan_very_2014} showed good results, while striking a balance between model complexity and computational cost. 
Our architecture is depicted in Figure~\ref{fig:vgg16network}. 

All three embedding generators were trained for 1400 epochs with a batch size of 32 and a learning rate of 1E-6 using the Adam optimizer. 
Before training, we split our data into a train (90\%) and validation (10\%) set. 
No test set was made since evaluating the classification task was not the focus of this study. 
We used the models with the best validation accuracy as embedding generators. 
We generated audio embeddings by performing a forward pass through the network with Mel-scaled spectrograms as input. 
The output vector of the 30 units of the pre-softmax embedding layer were taken as the audio embedding.

\subsection{Topic segmentation} 
\label{sect:meth-topicsegeval}

We evaluate our segmenter on a dataset of randomly concatenated hand-cut radio fragments from BNR Nieuwsradio, which we construct using a process as proposed by Choi et al.~\cite{choi_advances_2000}, who concatenate textual news snippets into a pseudo news bulletin for text segmentation. 
We concatenate audio fragments and their corresponding word-level transcripts into 60 minute sequences, and retain the topic boundaries as ground truth for training and evaluating the segmenter. 

We formalize the topic segmentation task as follows: 
given a sequence of words $W = [w_{1}, w_{2}, \ldots, w_{n}]$, the task is to predict the word-level topic boundaries $Y \in \{0,1\}^m$ where $m$ is the index of the word in the sequence, and $y_m = 1$ denotes that $w_m$ is a topic boundary.
As input we take a sequence of words $W$ of length $m$, represented by matrix $X \in \mathbb{R}^{m \times n}$ for $m$ words and $n$ features. 
For each word, we generate corresponding 30-dimensional audio embeddings by zero-padding the associated audio signal to 1 second, and passing it through our embedding generator described in Section~\ref{sect:meth-traingen}. 
We combine the 300-dimensional word embeddings $X_{w}$ with our audio embedding matrix $X_{a}$, yielding a set of combined feature vectors $X_{wa}$ per sequence. For example, the TXT+SEC task yields a feature vector $X_{wa}\in \mathbb{R}^{m \times (300+30)}$. 

Our topic segmenter is an RNN with LSTM units as proposed by Sheikh~\cite{sheikh_topic_2017}.
We train one model for each combination of features. 
We perform a hyper-parameter search in a predefined search space to find the best performing model for each combination of input features. 

We measure topic segmentation performance with the WinPR@k metric as proposed by Scaiano \cite{scaiano_getting_2012} and set $k=10$ for evaluation.
Evaluating segmentation at the exact (word) boundary-level is often deemed overly harsh; different systems may all score 0\% accuracy for missing exact boundaries, even if they show meaningful differences in segmentation output. 
For this reason, the metric treats near misses as hits, as controlled by $k$.

\section{Results}

\begin{table}[t]
  \small
  \centering
  \begin{tabularx}{\columnwidth}{l X X X}
    \hline
    \textbf{Task} & \textbf{No. of classes} & \textbf{\begin{tabular}[c]{@{}l@{}}Training\\accuracy\end{tabular}} & \textbf{\begin{tabular}[c]{@{}l@{}}Validation\\accuracy\end{tabular}} \\ \hline
    SEC & 50 & 46.9\% & 58.4\% \\
    FPC & 4 & 94.8\% & 73.0\% \\
    WC & 300 & 49.2\% & 58.0\% \\ \hline
    \end{tabularx}
  \caption{Accuracies on classification tasks of embedding generators}
  \label{table:clf-acc}
\end{table}

\subsection{Embedding generator classification accuracy}
In Table~\ref{table:clf-acc} we show the classification accuracy of different embedding generators.
The FPC task results in the highest validation accuracy score, at 73.0\%. 
For FPC, the training accuracy (94.8\%) is much higher than validation accuracy (73.0\%), suggesting that some form of overfitting occurred. 
Although the accuracy scores may be considered low, they show the models have learned to distinguish between samples to a reasonably degree. 
This implies these networks have learned representations that may be useful for extracting discriminating features from audio. 

\subsection{Topic segmentation}
Table \ref{table:results} summarizes the results for the down-stream topic segmentation task, using different embeddings as input.
We consider the performance of the text-only model as the baseline for our system. 
We find that our text-only baseline has relatively high precision (at 85.4\%) with lower recall (48.1\%).
The audio-only SEC outperforms this baseline with a 9.4\% increase in F1-measure, mostly attributed to improved recall.

Combining text with SEC results in the overall best performing model, with an uplift of 32.3\% in F1-measure over the baseline, with both precision and recall improving substantially.
More generally, combining text and audio embeddings results in an increased F1-measure in all cases, attributed to a substantial improvement in recall. 
Combining text with all audio embeddings (ALL, which combines TXT with SEC, FPC \& WC) shows a comparable precision but has a lower recall than the best model found.

\begin{table}[t]
  \small
  \centering
  \begin{tabularx}{\columnwidth}{c l X X X l}
  \hline
  & \textbf{Method} & \textbf{P} & \textbf{R} & \textbf{F1} & \textbf{Impr.} \\ 
  \hline
  baseline & TXT & 0.854 & 0.481 & 0.615 \\ 
  \hline
  audio    & SEC & 0.743 & 0.615 & 0.673 & +9.4\% ** \\ 
           & FPC & 0.584 & 0.592 & 0.588 & -4.4\% * \\ 
  \hline
  text     & \textbf{TXT+SEC} & \textbf{0.874} & \textbf{0.761} & \textbf{0.813} & \textbf{+32.3\% **} \\ 
  \&       & TXT+FPC & 0.787 & 0.697 & 0.739 & +20.2\% ** \\ 
  audio    & TXT+WC & 0.782 & 0.517 & 0.656 & +6.7\% ** \\ 
           & ALL & 0.864 & 0.760 & 0.809 & +31.5\% ** \\ 
  \hline
  \end{tabularx}
  \caption{Precision, Recall and F1-measures for WinPR@10. \\
  $*$/$**$ denotes statistical significance over baseline with p-value $<0.02$ / $\ll0.01$.}
  \label{table:results}
  \end{table}

We calculated statistical significance of the reported F1-score improvements by doing the non-parametric Friedman Aligned Ranks omnibus test, which confirmed there is a statistically significant difference between the methods.
Next, we compared each system to the baseline using a post-hoc Bonferroni-Dunn test~\cite{carterette_multiple_2012}. 
We found statistically significant differences from the baseline in all cases. 

In summary, we find that 
audio embeddings can be used to improve automatic, topically coherent segmentation of radio shows (\ref{rq1}). 
In addition, we confirm that the observed difference in segmentation performance between different audio embeddings shows that the specific classification task is of importance for the effectiveness of the embeddings for topic segmentation (\ref{rq2}). 
\section{Discussion \& Analysis}
In this study we explore the use of audio embeddings in a segmentation task of radio shows.
We show that audio embeddings improve the segmentation performance compared to text-only segmentation and that the specific training task of the embedding generator matters.

Our findings are consistent with the findings of T{\"u}r~\cite{tur_integrating_2001} and Tsunoo~\cite{tsunoo_hierarchical_2017} who observe similar effects with the inclusion of signal statistics and hand-crafted features. 
Our work focuses on the effectiveness of audio embeddings measured in the down-stream topic segmentation task and not on a comparison of audio features to other types of features, such as signal statistics or hand-crafted features. 
We do think that such a comparison could give valuable insights in the added value of audio embeddings in comparison to other features. 
Also in line with T{\"u}r~\cite{tur_integrating_2001} we see that segmentation based solely on audio features has comparable performance to (FPC) or even outperform (SEC) segmentation based on text alone. 
Kiela~\cite{kiela_learning_2017} concludes that audio embeddings with higher accuracy on the training task do not necessarily lead to better audio embeddings for the down-stream task. 
We find similar results with our SEC embeddings yielding a lower validation accuracy on the training task than the FPC embeddings, but yielding better performance in topic segmentation. 
Nevertheless, we find it interesting to see that the FPC task by itself shows a relatively high accuracy score, which suggests that it is possible to find meaningful patterns in the beginning, middle and end parts of radio fragments. 
Further inspection suggests it is relatively easy to detect the start of fragments (up to 82\% accuracy), trivial to recognize jingles (up to 99\%), but distinguishing between the middle and end parts of fragments proves to be more difficult. 

The high performance of the embeddings from the SEC task suggests that embeddings trained on a diverse dataset perform best. 
The SEC dataset is the only ``out of domain'' dataset, i.e., the WC and FPC datasets were derived from the same underlying corpus we used for segmentation. 

A preliminary comparison of network architectures and hyper-parameters for the embedding generator network resulted in the choice of the current experimental setup.
The use of a neural network architecture from the field of image classification in a sound analysis task is common practice and shows good results \cite{kiela_learning_2017, piczak_environmental_2015}.
Due to computational and time constraints we chose not to optimize this further when training specific embedding generators. 
Similar to Bengio~\cite{bengio_word_2014} and Kiela~\cite{kiela_learning_2017}, we used the pre-final fully connected layer of our CNNs as the embedding layer. 
We did not study the effect of using earlier layers or embedding layers of a different dimension. 

The main focus of this study was to evaluate the usefulness of audio embeddings in the down-stream task of topic segmentation, but other evaluation methods exist.
Schnabel~\cite{schnabel_evaluation_2015} discusses methods for evaluation of embeddings and makes a distinction between intrinsic and extrinsic evaluation methods. 
Using a down-stream task for performance evaluation is an example of such an extrinsic method.
However, intrinsic methods, such as clustering of the embeddings, are also possible and might provide insight in what is captured within the embedding space.
When clustering word embeddings, one could inspect the clusters of words and their meanings, but unlike words, sounds are not discrete objects and are therefore inherently more difficult to compare. 
To gain more insight in what is captured in our generated embeddings, we conducted experiments for relatedness and clustering. 
We explored whether the embeddings allow for emotion clustering, using the Ryerson Audio-Visual Database of Emotional Speech and Song (RAVDESS) dataset~\cite{livingstone_ryerson_2018}), which contains recordings of the same sentence spoken by different actors expressing different emotions. 
When conveying emotion by speech, acoustic properties of the speech differ~\cite{yang_linking_2001}. 
We expect that embeddings that captured acoustic properties of speech should be able to separate different emotions to some degree. 
However, none of the three embedding spaces were able to effectively cluster different emotions, suggesting that the embedding generators did not learn features specifically related to acoustic properties of speech.

We processed new, unseen radio shows into audio embeddings and clustered these embeddings. 
The t-SNE plot of SEC embeddings showed some formation of cluster regions. 
We developed an interactive application to study if audible patterns could be discerned in the cluster regions visible in the t-SNE plots of these radio show embeddings, but without any clear result. 

Evaluation of the down-stream task was performed on a dataset consisting of hand-cut radio fragments, which provides a clean dataset without 'noise' such as jingles or advertisement breaks. 
However, including this 'noise' might be beneficial for topic segmentation on real-aired radioshows, as it might contain information about topic boundaries.
\section{Conclusion}

In this paper we studied the use of audio embeddings for audio representation in a topic segmentation task.
Several studies show that using acoustic features in a segmentation task improves the performance, but none considered the use of audio embeddings for this.

We studied if audio embeddings can be used to improve the performance of a segmentation system and study the effect of different classification training tasks for audio embedding generators on the segmentation system.
We trained three audio embedding generators based on a multi-class classification task using three datasets from different domains: a non-speech sound event classification task, a part of radio show fragment classification task and a word classification task.
We use the pre-final fully connected layer of a Convolutional Neural Network based on the VGG16 architecture with Mel-scaled spectrograms as input and take output of the pre-final layer of the network as the audio embedding.
We evaluate performance of the embeddings using a dataset of artificially generated radio shows, based on hand-cut fragments. 

We show that audio embeddings generated from the non-speech sound event classification task significantly outperform our text-only baseline by 32.3\% in F1-measure. 
Combining audio embeddings with text features in all cases leads to a significant performance improvement, mostly by increasing the system's recall. 
A segmentation system that uses audio embeddings alone shows a comparable performance to using a text-only segmentation system.
We observe that different training tasks for the audio embedding generators lead to different performance of segmentation, and conclude that using audio embeddings can improve overall topic 
segmentation performance.

\subsubsection*{Acknowledgements}
The authors would like to thank Mahsasadat Shahshahani for her insights and guidance.
This work was supported by the University of Amsterdam Information Studies programme.

\vfill\pagebreak

\bibliographystyle{IEEEbib}
\bibliography{refs_clean}

\end{document}